\def\year{2022}\relax
\documentclass[letterpaper]{article} 
\usepackage{aaai22}  
\usepackage{times}  
\usepackage{helvet}  
\usepackage{courier}  
\usepackage[hyphens]{url}  
\usepackage{graphicx} 
\usepackage{natbib}  
\usepackage{caption} 
\DeclareCaptionStyle{ruled}{labelfont=normalfont,labelsep=colon,strut=off} 
\usepackage{algorithm}
\usepackage{algorithmic}

\usepackage{amsmath}
\usepackage{amsfonts}
\usepackage{amsthm}
\usepackage{arydshln}
\usepackage{dirtytalk}
\usepackage{booktabs}
\usepackage{multirow}

\usepackage{BOONDOX-cal}

\newcommand{\dnn}{\textbf{R}obust \textbf{I}nverse \textbf{D}esign under \textbf{Noise}}
\newcommand{\dab}{RID-Noise}

\newcommand{\bx}{\mathbf{x}}
\newcommand{\be}{\mathbf{\epsilon}}
\newcommand{\by}{\mathbf{y}}
\newcommand{\bz}{\mathbf{z}}
\newcommand{\yt}{\mathbf{y}_{\textbf{target}}}
\newcommand{\bF}{\mathbcal{F}}
\newcommand{\bR}{\mathbcal{R}}
\newcommand{\bB}{\mathbcal{B}}

\newcommand{\bE}{\mathbb{E}}
\newcommand{\fforward}{f_{\text{forward}}}

\newcommand{\refeq}[1]{Eq. (\ref{#1})}
\newcommand{\reffig}[1]{Figure \ref{#1}}

\newcommand{\refalg}[1]{Algorithm \ref{#1}}
\newcommand{\reftab}[1]{Table \ref{#1}}
\newcommand{\refprop}[1]{Property \ref{#1}}

\newtheorem{theorem}{Theorem}

\newtheorem{property}[theorem]{Property}

\usepackage{newfloat}
\usepackage{listings}
\floatstyle{ruled}
\newfloat{listing}{tb}{lst}{}
\floatname{listing}{Listing}

\title{RID-Noise: Towards Robust Inverse Design \\
under Noisy Environments 
\footnote{Jia-Qi Yang is the corresponding author. 
This work was supported by NSFC61773198, 
Recruitment Program for Young Professionals and Fundamental Research Funds for the Central Universities.}}
\author {
        Jia-Qi Yang\textsuperscript{\rm 1},
        Ke-Bin Fan\textsuperscript{\rm 2},
        Hao Ma\textsuperscript{\rm 2},
        De-Chuan Zhan\textsuperscript{\rm 1}
}
\affiliations {
    \textsuperscript{\rm 1} State Key Laboratory for Novel Software Technology, Nanjing University, Nanjing, China \\
    \textsuperscript{\rm 2} Research Institute of Superconductor Electronics (RISE) \\
    School of Electronic Science and Engineering, Nanjing University, Nanjing, China \\
    \textsuperscript{\rm 1}\{yangjq,zhandc\}@lamda.nju.edu.cn\\
    \textsuperscript{\rm 2} kebin.fan@nju.edu.cn, mf20230071@smail.nju.edu.cn
}

\begin{document}

\maketitle

\begin{abstract}
    From an engineering perspective, a design should not only perform well in an ideal condition,
    but should also resist noises.
    Such a design methodology, namely robust design,  has been widely implemented in the industry for product quality control.
    However, classic robust design requires a lot of evaluations
    for a single design target,
    while the results of these evaluations could not be reused for a new target.
    To achieve a data-efficient robust design, we propose \dnn{} (\textbf{\dab{}}),
    which can utilize existing noisy data to train a conditional invertible neural network (cINN).
    Specifically, we estimate the robustness of a design parameter by its predictability,
    measured by the prediction error of a forward neural network.
    We also define a sample-wise weight, which can be used in the maximum
    weighted likelihood estimation of an inverse model based on a cINN.
    With the visual results from experiments, we clearly justify
    how \dab{} works by learning the distribution and robustness from data.
    Further experiments on several real-world benchmark tasks with noises 
    confirm that our method is more effective than other state-of-the-art inverse design methods.
    Code and supplementary is publicly available at \textit{https://github.com/ThyrixYang/rid-noise-aaai22}
\end{abstract}

\section{Introduction}

\textit{Robust design has been developed with the expectation that an insensitive design can be obtained.
    That is, a product designed by robust design should be insensitive to external noises or tolerances.
    An insensitive design has more probability to obtain a target value, although there are uncertain
    noises.}~\cite{robust_design_overview}

\begin{figure}
    \centering
    \includegraphics[width=\linewidth]{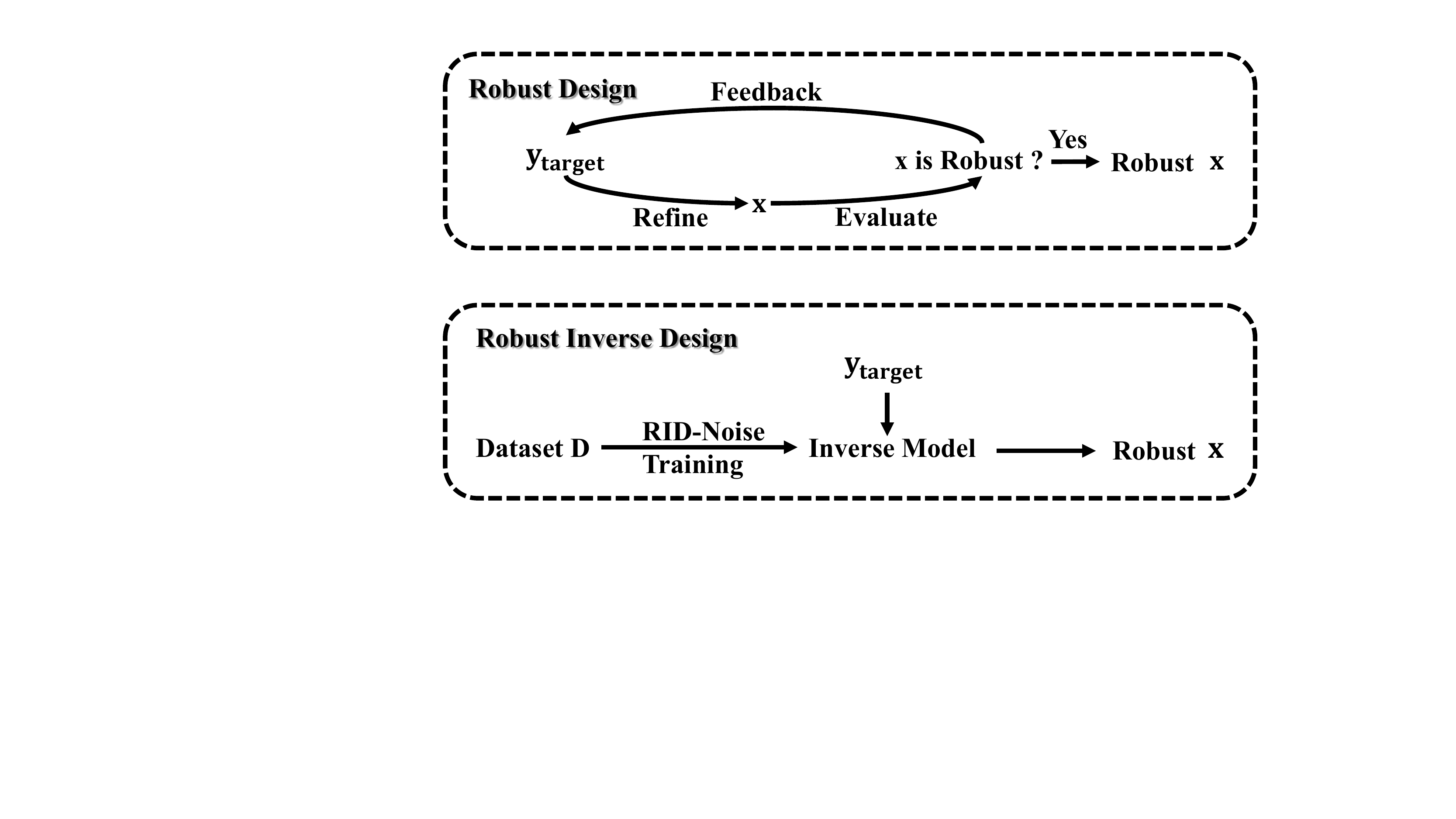}
    \caption{The classic \textbf{robust design} methodology requires many
    evaluate-feedback loops to evaluate and improve
    the robustness of a design parameter $\bx$ towards a given design target $\textbf{y}_{\textbf{target}}$.
    We propose the \textbf{robust inverse design} setting and
    the \textbf{\dab{}} method that can learn the design parameter distributions
    as well as robustness from a dataset $D$, which can be collected during past design tasks for different targets.
    During inference, \dab{} can output robust candidate $\bx$
    for a given target $\textbf{y}_{\textbf{target}}$ without further evaluation.
    }
    \label{fig:intro_rid}
\end{figure}

Pioneered by ~\citet{taguchi}, \textbf{robust design}
nowadays has become an essential design concept in industry engineering~\cite{prin_robust_design}.
In a robust design task,
the objective is to find a design parameter $\bx$ that can stably achieve a design target $\yt$ under various noises.
For example, assuming a ball is moving along a potential energy surface,
the corresponding design target is to make the velocity of the ball $\yt=0$,
and the design parameter $\bx$ is the position to put the ball.
Typically, the ball can stand still both on a local minimum $\bx_1$ or on a saddle point $\bx_2$ in ideal conditions.
However, the ball on a saddle point can easily slide down with a tiny perturbation, resulting in a large velocity $\yt \gg 0$;
on the contrary, such a noise is less likely to accelerate a ball lying on a local minimum.
Thus, $\bx_1$ is more robust compared to $\bx_2$.
The Hessian matrix of the surface can be used as a criterion of robustness in this example.
The robust design routine builds upon an evaluate-feedback loop to improve the robustness, as depicted in the upper panel
of \reffig{fig:intro_rid}.
Therefore, the classic robust design methods are evaluation intensive since optimizing robustness requires a large number of
iterations, and the data generated from these evaluations could not be reused for a new target ~\cite{experiment_complex}.

To improve data efficiency, a paradigm known as \textbf{inverse design} has been proposed, in which
the inverse process emphasizes that no feedback loop is required for a new target,
since the inverse design $\by \rightarrow \bx$ is achieved directly.
For example, the Tandem method ~\cite{tandem} trains a deterministic encoder-decoder structure in which the encoder can be used to
produce a design parameter for a given target. ~\citet{analyzing_inverse_problem} proposed to use invertible neural networks to
learn a generative model that can create several different candidates.
These machine learning approaches are preferable for better data efficiency ---
once an accurate model is trained, no more evaluations are needed for a new design target.
However, these methods cannot be adopted to robust design problems
since the training data is assumed to be sampled from a deterministic environment.

The lower panel of \reffig{fig:intro_rid} describes the concept of a general \textbf{robust inverse design} problem,
which aims at data-efficient robust design.
There are two notable challenges in robust inverse design problems.
First, the dataset is collected from a noisy environment,
so that a robust inverse design method should be able to avoid the noisiest data points.
Secondly, the training data distribution contains expert knowledge of the way to explore the manifolds of feasible
parameters for a given target. Thus, a robust inverse design method should learn this distribution
while ensuring robustness.

\begin{figure}
    \centering
    \includegraphics[width=\linewidth]{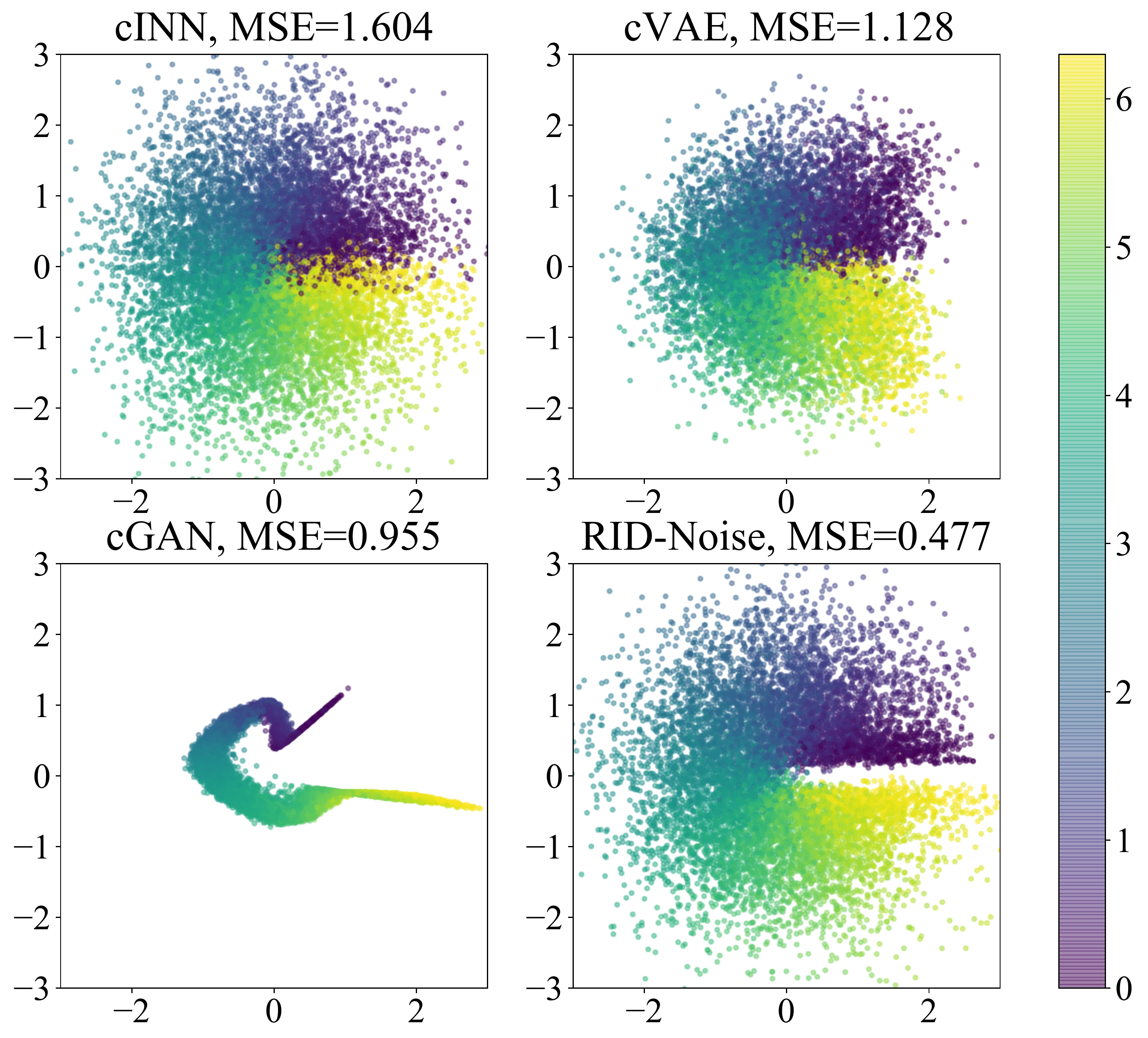}
    \caption{The value of $y$ is the radian of each point, which is denoted in color.
        Stochasticity is introduced by adding a random noise
        to the design parameter $\bx \in \mathbb{R}^2$ (visualized as two axes).
        The area near $y=0$ is special since a small perturbation of $\bx$ will change $y=0$ to $y=2\pi$ abruptly, thus not robust.
        The proposed \dab{} method can learn to keep away from this dangerous area, while exploring the remaining
        parameter space evenly.
    }
    \label{fig:intro_cart}
\end{figure}

In this work, we propose a \dnn{} (\textbf{\dab{}}) method to tackle inverse design problems which demand robustness.
In our method, we estimate the robustness of each training data point based on the predictability,
which is achieved by measuring the cross-validation error of a feed-forward neural network.
Also, we propose to train normalizing flows~\cite{normalizing_flow_review} with weighted likelihood,
where the balance between density estimation and robustness can be controlled by a single hyper-parameter $\tau$.
We further show that \dab{} can work empirically by learning the distribution and robustness from data,
as visualized in \reffig{fig:intro_cart}.
With thorough investigations on three benchmark tasks, we demonstrate the effectiveness of our method.
And the experimental results confirm that our method consistently outperforms state-of-the-art inverse design methods.
Our main contributions can be summarized as follows:

\begin{itemize}
    \item{
                We propose a robust inverse design benchmark, which aims to utilize existing data to
                train an inverse model capable of generalizing to new design targets.
          }

    \item{We develop a robust inverse design method via training normalizing flows with weighted likelihood loss,
                which can strike a balance between density estimation and robustness.
          }

    \item{We design several experiments that visually show when the proposed \dab{} method works 
                and why it should work. The results on three classic benchmark tasks confirm the effectiveness of the proposed method.
          }

\end{itemize}

\section{Related Work}

\subsection{Robust Design}

In real-world engineering design problems,
unexpected deviations from the initial planned values are ubiquitous because of
inevitable noises or uncontrollable variables in engineering processes~\cite{taguchi_robust_engineering}.
The Taguchi method ~\cite{taguchi} uses a quadratic loss to measure the deviation from a design target $\yt$,
and utilizes the orthogonal arrays for experiment designs. However, to form the orthogonal arrays,
the design parameters $\bx$ are required to be discrete, 
and the number of experiments increases exponentially as the level of parameters raises, 
which is costly~\cite{experiment_complex}.
Therefore, optimization techniques were introduced to tackle
with continuous parameter space and
constraints, e.g. robust optimization ~\cite{robust_optimization}. 
Such a method as depicted in the upper panel of \reffig{fig:intro_rid} requires
to estimate the robustness and improve the objective iteratively,
causing the evaluations during inference still expensive. 

\subsection{Machine Learning Aided Design}

Machine learning techniques for engineering designs have
attracted much attention 
in recent years ~\cite{ml_spotlight,inverse_design_chem_rl,mm_nature_photonics}.
One direction of machine learning aided design is to train a surrogate model,
so that the experiments can be simulated using the model efficiently~\cite{deep_neural_network_nature,robust_design_icml}.
For example, ~\citet{mm_task} trained a deep neural network to model complex all-dielectric metasurface systems,
and performed fast forward dictionary search to achieve inverse design.
However, these methods cannot be adopted to robust design since it is challenging to train a surrogate model of a noisy environment.
Another direction is to train a generative model to directly create design
parameters with desired properties ~\cite{benchmark_2018,tandem,na_journal,deep_inverse_design_nano}. 
For example, ~\citet{analyzing_inverse_problem} proposed to train an invertible neural
network model to learn an inverse mapping from targets to design parameters.
However, these methods assume the environment is deterministic, thus could not be adopted to robust design problems.

\section{Proposed Method}

\subsection{Background}

In a robust design problem, the forwarding process is normally not deterministic.
We denote the forwarding process as a conditional distribution
$p(\mathbf{y}|\mathbf{x})$, where $\mathbf{x}\in \mathbb{R}^{d_x}$ is a vector of the
design parameters, and $\mathbf{y} \in \mathbb{R}^{d_y}$ is a vector which contains the information of the desired properties, 
for example, the transmission spectrum of a photonic device ~\cite{benchmark_2020}.
Given a target $\yt$, the robustness objective of a parameter $\bx'$ can be defined by the expected loss
\begin{equation}\label{rid_objective}
  \mathbcal{L}_{p(\by|\bx)}(\bx', \yt) = \bE_{\by'\sim p(\mathbf{y}|\mathbf{x}')} \mathbcal{l}(\by', \yt)
\end{equation}
The robust inverse design problem can be defined by minimizing the robustness objective
\begin{equation}\label{rid_optimization}
  \min_{\bx'} \mathbcal{L}_{p(\by|\bx)}(\bx', \yt)
\end{equation}
Where $\mathbcal{l}(\by, \yt)$ is a loss function that measures the difference between $\yt$ and $\by$.
In the following context, we assume the loss function is the mean squared error (MSE)
defined by $\mathbcal{l}(\by, \yt)=||\by - \yt||_2^2$,
which is a typical choice in robust design known as quadratic loss function ~\cite{robust_design_overview}.

As mentioned above, a general robust design procedure requires many experiments to try on feasible input $\bx$
to minimize the expected loss \refeq{rid_objective}, which can be viewed as an online optimization process of the target $\yt$.
This procedure is depicted in the upper panel of \reffig{fig:intro_rid}.
For a new design target, the entire online procedure has to start from the beginning again, 
whereas those data from previous experiments are abandoned. 
To use all these data efficiently, instead,
we consider an offline learning setting with an available dataset
$D=\{(\mathbf{x}_i, \mathbf{y}_i), 1 \le i \le n\}$, where $\mathbf{x}_i$ is sampled from
a distribution $\mathbf{x}_i \sim p(\mathbf{x})$. Then $\by_i$ is generated by a forwarding process
$\mathbf{y}_i \sim p(\mathbf{y}|\mathbf{x}_i)$, and the joint distribution is denoted by
$p(\mathbf{x}, \mathbf{y})=p(\mathbf{x})p(\mathbf{y}|\mathbf{x})$.
Usually, we only have the dataset $D$ without knowing the explicit form of $p(\bx)$.
The dataset is collected by experts with domain knowledge and design experience.

\subsection{Measuring the Robustness with Dataset}

To achieve offline training of robust inverse design, a crucial problem arises:
how to measure the robustness of a design parameter $\bx$? 
Our main idea is that \textbf{robustness can be interpreted by predictability}. 
That is, if we can predict the response of $\bx$ accurately, $\bx$ is believed to be a robust parameter, since the outcome of 
$\bx$ is nearly deterministic.
Based on this intuition, we propose to train a forward model $\fforward(\bx)$ on training dataset, 
and predict the corresponding $\by$ on validation dataset. 
The prediction error is used to measure a sample-wise predictability for $\bx_i$ as in \refeq{sample_robustness}.

\begin{equation}\label{sample_robustness}
  \mathbcal{r}(\bx_i) = \mathbcal{l}(\by_i, \fforward(\bx_i))
\end{equation}

The $\mathbcal{r}(\bx_i)$ can be estimated for every $(\bx_i, \by_i)$ in the whole dataset with cross-validations.
We analyze the relationship between this sample-wise predictability and 
the objective of robust design defined in \refeq{rid_objective} in the following sections.

\subsubsection{Estimating the Robustness}
A straightforward way is to estimate the expected loss in \refeq{rid_objective} by Monte Carlo method:
\begin{equation}
  \mathbcal{L}_{p(\by|\bx)}(\bx', \yt) \approx \frac{1}{N} \sum_{i=1}^{N} \mathbcal{l}(\by_i, \yt)
\end{equation}
Where $\by_i$ is experimentally drawn from $p(\by|\bx')$.
This is the core idea of robust design as an experiment design method.
However, it is impossible to use in an offline training since the target $\yt$ is not defined yet during training.

Therefore, to estimate the robustness, our first step is to define a robustness measure of $\bx'$ without regarding $\yt$, 
that is, we define a target agnostic robustness $\bR$ as: 
\begin{align}
  \bR_{p(\by|\bx)}(\bx') & = \mathbcal{L}_{p(\by|\bx)}(\bx', \bF_{p(\by|\bx)}(\bx'))  \label{ta_robustness} \\
  \bF_{p(\by|\bx)}(\bx') & = \bE_{\by' \sim p(\by|\bx')} \by'  \label{func_mean}
\end{align}
We use the function $\bF(\bx)$ to denote the mean of response $\by$ given $\bx$.
The target agnostic robustness defined in \refeq{ta_robustness} measures the expected deviation from the
mean value of the forwarding process defined in \refeq{func_mean}.

\begin{property}\label{robust_decomp}
  With the MSE loss
  \begin{align}
    \mathbcal{L}_{p(\by|\bx)}(\bx', \yt) & = \bR_{p(\by|\bx)}(\bx') + \mathbcal{B}(\bx', \yt)\label{bvd} \\
    \mathbcal{B}(\bx', \yt)              & =  ||\bF_{p(\by|\bx)}(\bx') - \yt||_2^2 \label{bias}
  \end{align}
\end{property}
\noindent \textit{Proof Sketch}:
  \refeq{bvd} is a different view of the bias-variance decomposition ~\cite{prml}. 
  A detailed proof is in the supplementary material.

The \refprop{robust_decomp} indicates that the robust design objective \refeq{rid_objective} can be decomposed into 
the sum of a target agnostic robustness $\bR$ and a bias term $\bB$.

\begin{property}\label{n2n}
  Training a forward model $\fforward(\bx)$ with the MSE loss converges to the mean function $\bF(\bx)$.
\end{property}
The \refprop{n2n} has been successfully applied to training neural networks for image denoising ~\cite{noise2noise}. 
Utilizing \refprop{n2n}, we train a feed-forward neural network $\fforward(\bx)$ to estimate the mean function $\bF(\bx)$.

\subsubsection{Sample-Wise Robustness}

The target agnostic robustness $\bR$ defined in \refeq{ta_robustness} still requires sampling. However,
since $\bF$ can be estimated by $\fforward$, 
the sample-wise predictability \refeq{sample_robustness} can be viewed as an estimation of \refeq{ta_robustness}: 
\begin{align}
  \bE_{\by_i \sim p(\by|\bx_i)} \mathbcal{r}(\bx_i) &= \bE_{\by_i \sim p(\by|\bx_i)} \mathbcal{l}(\by_i, \fforward(\bx_i)) \\
                        &\approx \bE_{\by_i \sim p(\by|\bx_i)} \mathbcal{l}(\by_i, \bF(\bx_i)) \\
                        & = \bR_{p(\by|\bx_i)}(\bx_i)
\end{align}
Intuitively, if a data point $(\bx_i, \by_i)$ is near $\bF(\bx_i)$, the chance that $\bx_i$ is robust will be higher. 
Thus we define \refeq{sample_robustness} as a sample-wise robustness measure, 
which is exactly the predictability defined before.

\subsubsection{Tackling the Bias Term}

The sample-wise robustness $\mathbcal{r}(\bx_i)$ defined above measures the target agnostic robustness $\bR$.
To minimize the bias term $\bB$, we can re-label the dataset $(\bx_i, \by_i)$ to $(\bx_i, \bF(\bx_i))$. 
Empirically, we find that using the original $\by_i$ leads to similar results.
Intuitively, a robust $\bx_i$ will generate a $\by_i$ close to its expected value, 
while the $\bx_i$ that is not robust will be down-weighted so that the value $\by_i$ has little effect.
Thus we use the original $\by_i$ in experiments.

\subsection{Generative Inverse Design with cINNs}

There are several merits using a generative model to learn $p(\bx|\by)$ instead of learning a deterministic mapping 
(such as the Tandem method ~\cite{tandem}).
(1) In a nondeterministic design problem, given a $\by$, the corresponding $\bx$ is usually not unique.
A generative model is able to map
one $\by$ to many $\bx$, which is more reasonable in the inverse design task.
This advantage can be used to guide the experiment process, for example, experts can select some of them to conduct experiments.
(2) The existing data distribution can be viewed as a prior information that contains some expert knowledge about
how to explore the design parameter space. Thus, a generative model can utilize these knowledge
by learning this distribution $p(\bx|\by)$ directly.

The conditional invertible neural networks (cINNs) were proposed by ~\citet{cinn} as an extension of INNs.
We choose cINNs to learn $p(\bx|\by)$ from data.
The cINNs are normalizing flows implemented with neural networks, which model a generative process as a sequence of
invertible transformations implemented by coupling blocks~\cite{real_nvp, glow_inn}. 
The main idea of flow-based modeling is to express $\mathbf{x}$ as an
invertible transformation $\mathbf{x}=f(\mathbf{z}; \by, \theta)$ of a
real vector $\mathbf{z}$ sampled from $p_\mathbf{z}(\mathbf{z})$:
\begin{equation}
  \mathbf{x}=f(\mathbf{z}; \by, \theta) \quad \text { where } \quad \mathbf{z} \sim p_{\mathrm{z}}(\mathbf{z})
\end{equation}
Where $\by$ is the conditioning variable and $\theta$ is the trainable parameters of the cINN.
The distribution $p(\mathbf{z})$ is a known distribution which is easy to analyze, usually gaussian.
The core of normalizing flows is the change of variables formula~\cite{normalizing_flow_review}:
\begin{equation}
  p_{\mathrm{x}}(\mathbf{x})=p_{\mathrm{z}}(\mathbf{z})\left|\operatorname{det} J_{\mathbf{z}}\right|^{-1} \quad \text
   { where }  \mathbf{z} = f^{-1}(\mathbf{x}; \by, \theta)
\end{equation}
$J_{\mathbf{z}}$ is the Jacobian given by
\begin{equation}\label{jacobian}
  J_{\mathbf{z}}=\left.\frac{\partial f(\mathbf{z}; \by, \theta)}{\partial \mathbf{z}}\right|_{\mathbf{z}=f^{-1}(\mathbf{x}; \by, \theta)}
\end{equation}

The cINNs can calculate the Jacobian \refeq{jacobian} efficiently. 
Consequently, with the cINNs, the likelihood of $p(\bx|\by)$ can be calculated analytically by the change of variable formula, which
is preferable comparing to other generative models such as generative adversarial networks (GANs) ~\cite{gan}
or variational auto-encoders (VAEs) ~\cite{vae}, in which the likelihood cannot be calculated efficiently. 
We discuss an alternative approach in the supplementary material.

\begin{algorithm}[tb]
  \caption{Estimating the \dab{} Weights}\label{alg:weight}
  \textbf{Input}: Dataset $D=\{(\bx_1, \by_1), (\bx_2, \by_2), ..., (\bx_n, \by_n)\}$ \\
  \textbf{Parameter}: Number of fold $k$; Weight temperature $\tau$\\
  \textbf{Output}: Weights $\mathbcal{w} \in \mathbb{R}^n$.
  \begin{algorithmic}[1] 
    \STATE Split dataset $D$ evenly into $k$ datasets $\{D_1, D_2, ..., D_k\}$.\label{alg:weight:cv1}
    \FOR{$i$ \textbf{in} $\{1,...,k\}$}
    \STATE Training dataset $D_{t}=\{D_j, j\ne i\}$.
    \STATE Validation dataset $D_{v}=D_i$.
    \STATE Train a model $\fforward$ to fit $\mathbcal{l}(\by, \fforward(\bx))$ on $D_t$.
    \STATE Calculate $\mathbcal{r}_{i}=\mathbcal{l}(\by, \fforward(\bx)), (\bx, \by) \in D_v$. \label{alg:weight:cv2}
    \ENDFOR
    \STATE $\mathbcal{r} = \{\mathbcal{r}_i, i \in \{1,...,k\}\}$.
    \STATE $\mathbcal{r} = \frac{\mathbcal{r}}{\textbf{mean}(\mathbcal{r})}$
    \STATE $\mathbcal{w} = \textbf{exp}(-\tau \mathbcal{r})$
    \STATE $\mathbcal{w} = \frac{\mathbcal{w}}{\textbf{mean}(\mathbcal{w})}$ + \text{eps}
    \STATE \textbf{return} $\mathbcal{w}$, where $\mathbcal{w}_i = \mathbcal{w}(\bx_i, \by_i, D)$
  \end{algorithmic}
\end{algorithm}
\subsection{Training Robust Design Generator}
The main idea of training a robust design generator is to consider the robustness in the training process
of the generative model. Here, we introduce the weighted likelihood method to train a
cINN model for the robust inverse design problem.

The weighted likelihood method has been used to conduct
robust bayesian inference ~\cite{robust_baysian_inference_wl, wl_intro}, or
local likelihood ~\cite{local_likelihood}.
Given a dataset $D=\{(\bx_1, \by_1), (\bx_2, \by_2), ... (\bx_n, \by_n)\}$, the weighted likelihood function is defined as
\begin{equation}
  L^w(D;\theta)= \Pi^{n}_{i=1} q(\bx_i| \by_i; \theta)^{\mathbcal{w}(\bx_i, \by_i, D)}
\end{equation}
Where $\mathbcal{w}(\bx_i, \by_i, D)$ is the weight function that gives a weight 
given $\bx_i$, $\by_i$ and all the available dataset $D$:
\begin{equation}
  \mathbcal{w}(\bx_i, \by_i, D) = \textbf{exp}\left(-\tau \mathbcal{r}(\bx_i)\right) \quad \textbf{where} \quad \tau \ge 0
\end{equation}
That is, the larger the robust loss $\mathbcal{r}$, the smaller the weight $\mathbcal{w}$.
The procedure of estimating $\mathbcal{w}$ is summarized in \refalg{alg:weight}. 
Line [1-7] estimates the sample-wise robustness $\mathbcal{r}$ via cross-validations. 
Line [8-9] normalizes the $\mathbcal{r}$ to make it invariant to the scale of $\by$.
Line [10-11] converts $\mathbcal{r}$ to weights $\mathbcal{w}$ required in the weight likelihood method.
Using the weights defined above, the weighted negative log-likelihood (WNLL) loss function is
\begin{align}
  \text{WNLL}(D; \theta) &= -\log L^w(D; \theta) \label{weighted_likelihood}\\
  & = -\sum^{n}_{i=1} w(\bx_i, \by_i, D) \log q(\bx_i| \by_i; \theta)
\end{align}
The WNLL can be trained end-to-end to optimize the parameters $\theta$ of the cINN model.

There are at least three advantages for training with the proposed WNLL loss. 
First, the weighted likelihood method connects robustness and density estimation explicitly,
by down-weighting the data points that are believed to be less robust.
Secondly, 
the tolerance for the lack of robustness is dependent on specific problems, but it can be easily 
adjusted with tuning the hyper-parameter $\tau$.
Thirdly, the WNLL loss will be unchanged for the distribution of data points with robustness
since the weights of these points are close to 1.
We further analyze these properties empirically in the following sections.

\section{Experiments}

\begin{table*}[t]
    \centering
    \begin{tabular}{llllllllll}
        \toprule
        \multirow{2}{*}{Methods} & \multicolumn{3}{c}{Kinematics} & \multicolumn{3}{c}{Balllistics} & \multicolumn{3}{c}{Meta Meterial}                                                                                                                               \\
        \cmidrule(lr){2-4}\cmidrule(lr){5-7}\cmidrule(lr){8-10}
                                 & $\mathcal{n}_x$                & $\mathcal{n}_y$                 & $\mathcal{n}_{xy}$                & $\mathcal{n}_x$    & $\mathcal{n}_y$    & $\mathcal{n}_{xy}$ & $\mathcal{n}_x$    & $\mathcal{n}_y$    & $\mathcal{n}_{xy}$ \\
        \midrule
        NA                       & 0.9144                         & 0.4910                          & 1.4256                            & 26.882             & 21.789             & 35.494             & \underline{0.0181} & \underline{0.0412} & \underline{0.0569} \\
        Tandem                   & 1.7025                         & 1.7084                          & 1.8169                            & 2.0727             & 2.1650             & 2.2600             & 0.0535             & 0.0890             & 0.0984             \\
        cVAE                     & 0.2252                         & 0.1884                          & 0.3897                            & 0.0707             & \underline{0.0966} & 0.2314             & 0.0249             & 0.0505             & 0.0739             \\
        cINN                     & \underline{0.1732}             & \underline{0.1240}              & \underline{0.2809}                & 0.0779             & 0.1062             & 0.1874             & 0.0226             & 0.0475             & 0.0708             \\
        INN-LL                   & 0.1984                         & 0.2730                          & 0.3748                            & 1.9229             & 2.1297             & 2.2232             & 0.0325             & 0.0805             & 0.1026             \\
        INN-MMD                  & 1.8678                         & 1.5818                          & 1.9078                            & 1.8451             & 1.5361             & 1.1550             & 0.0545             & 0.0992             & 0.1054             \\
        cGAN                     & 0.2186                         & 0.1374                          & 0.3209                            & \underline{0.0654} & 0.1114             & \underline{0.1745} & 0.0300             & 0.0566             & 0.0785             \\
        \textbf{\dab{}}          & \textbf{0.0874*}               & \textbf{0.0726*}                & \textbf{0.1711*}                  & \textbf{0.0235*}   & \textbf{0.0670*}   & \textbf{0.1072*}   & \textbf{0.0147*}   & \textbf{0.0409}    & \textbf{0.0516*}   \\
        \bottomrule
    \end{tabular}
    \caption{\label{tab:benchmark}Benchmark performance in mean squared error (MSE).
        The bold value marks the best one in one column,
        while the underlined value corresponds to the best one among all baselines.
        Here, * indicates statistical significant improvement compared to the best baseline measured by t-test at $p$-value of 0.05.}
\end{table*}

To evaluate the proposed model, we conduct a set of experiments to answer the following questions:

\noindent\textbf{Q1:} How does \dab{} perform on real-world inverse design problems compared to state-of-the-art methods?

\noindent\textbf{Q2:} Why does the proposed \dab{} method work on robust inverse design problems?

\noindent\textbf{Q3:} How does the hyper-parameter $\tau$ affect the performance of \dab{}?

\subsection{Compared Methods}

We compare our method with the state-of-the-art methods:

\begin{itemize}

    \item[--] \textbf{cVAE}: The conditional variational auto-encoder is a generative model which optimizes the
        evidence lower bound to maximize the conditional likelihood $p(\bx|\by)$ ~\cite{cvae},
        which can be adopted to inverse design problems straightforwardly ~\cite{benchmark_2018}.

    \item[--] \textbf{cINN}: Invertible neural networks (INNs) can be used to
        optimize the likelihood $p(\bx)$ directly ~\cite{real_nvp, glow_inn}.
        ~\citet{benchmark_2018} proposed to train conditional invertible neural networks (cINN) to optimize
        the conditional likelihood $p(\bx|\by)$ in inverse problems.

    \item[--] \textbf{INN-MMD}: ~\citet{analyzing_inverse_problem} proposed to train INNs
        with the cyclic maximum mean discrepancy (MMD) loss to learn a $p(\bx|\by, \bz)$ generator bi-directionally.

    \item[--] \textbf{INN-LL}: ~\citet{benchmark_2018} suggested to train INNs with log likelihood (LL) instead of MMD.

    \item[--] \textbf{Neural Adjoint (NA)} trains a forward surrogate model with a neural network,
        and the adjoint method (gradient descent with respect to input)
        can be conducted using the surrogate model \cite{benchmark_2020}.

    \item[--] \textbf{cGAN} trains a generator to generate novel samples from the conditional distribution,
        and a discriminator is trained simultaneously to distinguish generated and real samples ~\cite{gan, cgan}.
        ~\citet{inverse_design_gan} adopted cGAN to design metasurface of microscale devices.

    \item[--] \textbf{Tandem} trains a forward model ($\bx \rightarrow \by$), then a backward model is pre-pended to the forward model
        to learn a ($\by \rightarrow \bx$) mapping ~\cite{tandem}.

    \item[--] \textbf{\dab{}}: Our \dab{} method.

\end{itemize}

\subsection{Implementation and Parameter Settings}

For fair comparison, all hyper-parameters are tuned carefully for all methods.
We use the Adam optimizer with learning rate tuned within $[10^{-2}, 10^{-3}, 10^{-4}]$,
and weight decay tuned within $[10^{-4}, 10^{-5}, 10^{-6}, 10^{-7}]$ for all methods.
The network structures are tuned for each method, for example,
the MLP layer number and layer width in Tandem, cVAE, cGAN, and NA method;
type of coupling blocks of INN based methods such as INN-LL, INN-MMD, cINN;
the clamp value of the coupling blocks of INN based methods. There are also some hyper-parameters
for a specific method, such as the weights of bi-directional training in INN-MMD.
All these methods are tuned to make sure they are well behaved or their performance are well optimized.
We provide a more detailed summary in the supplementary material.

\begin{figure}
    \centering
    \includegraphics[width=\linewidth]{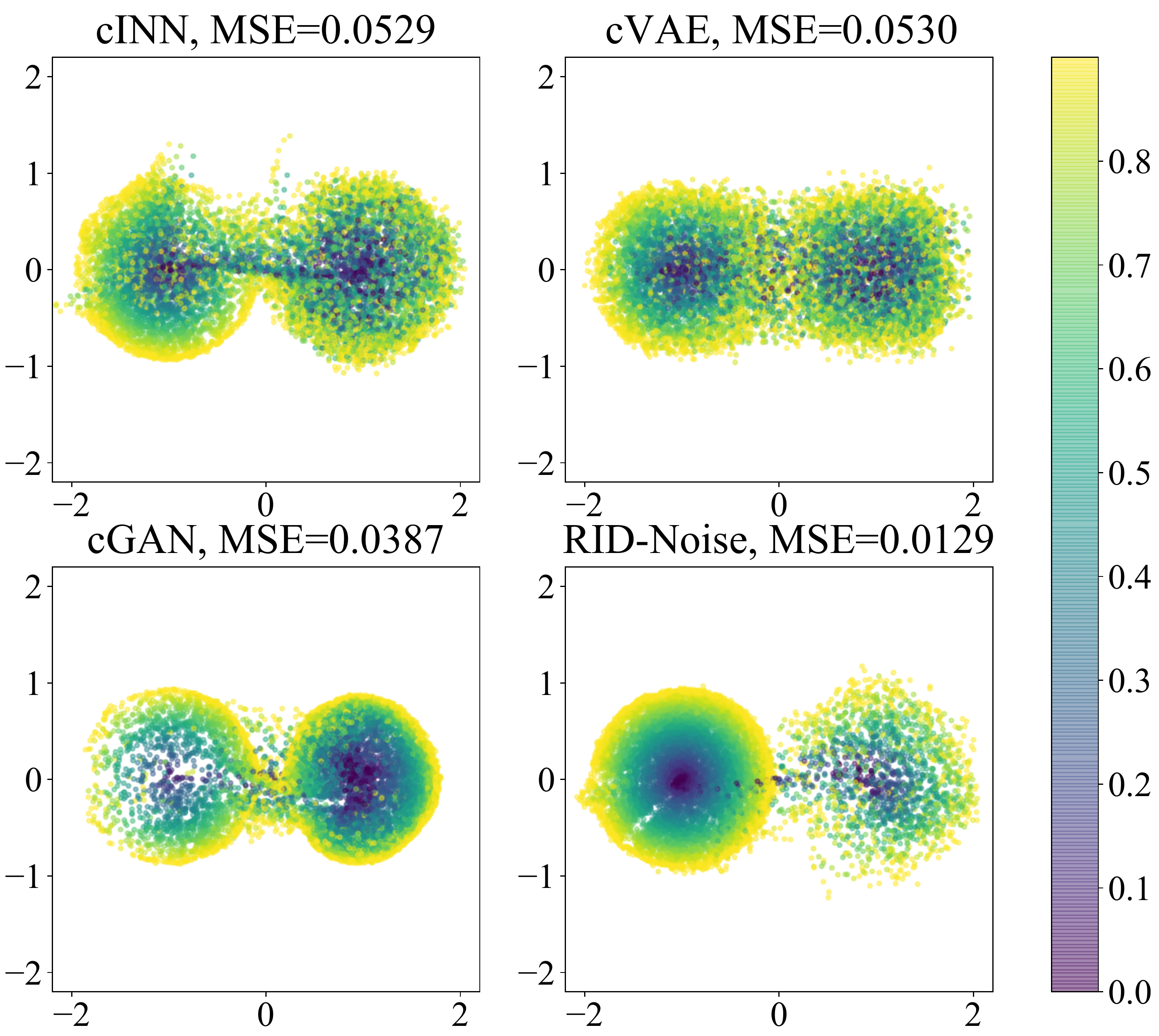}
    \caption{Results of generative methods on the Radius task.
        By construction, a robust inverse design method should learn to choose the $\bx$ on the left side.
        The cVAE and cINN methods struggle to fit the noisy data distribution, resulting in high expected error.
        The cGAN prefers the right side, which is not the desired behavior.
        This may be caused by that the discriminator cannot distinguish the noisy data and generated data on the right side.
        Only the \dab{} method successfully learns to generate robust $\bx$ on the left side.}
    \label{fig:two_cluster}
\end{figure}

\subsection{Simulating the Randomness in Environments}

Randomness exists naturally in real world, however,
the cost of testing is prohibitive.
To verify the effectiveness of the proposed method,
we select three benchmark tasks from previous research on inverse design:
\begin{itemize}
    \item[--] \textbf{Kinematics}: An articulated arm moves vertically along a rail and rotates at
        three joints, the inverse design problem is
        to find the angle parameters to achieve a given point ~\cite{analyzing_inverse_problem}.

    \item[--] \textbf{Ballistics}: A ball is thrown forward then lands on the ground,
        the inverse design problem is to find the angle, velocity and the position given
        the landing position ~\cite{benchmark_2018}.

    \item[--] \textbf{Meta-Material}: The goal of this task,
        is to design the radii and heights of four cylinders of a meta-material so that
        it produces a desired electromagnetic reflection spectrum ~\cite{benchmark_2020}.

\end{itemize}

All three tasks can be denoted by a deterministic forwarding function $g(\bx)=\by$,
where $\bx$ denotes the corresponding design parameters, and $\by$ is the response vector.
We construct stochastic problems by adding random noises to the input $\bx$ and output $\by$ of the function $g(\bx)=\by$.

\begin{figure}
    \centering
    \includegraphics[width=\linewidth]{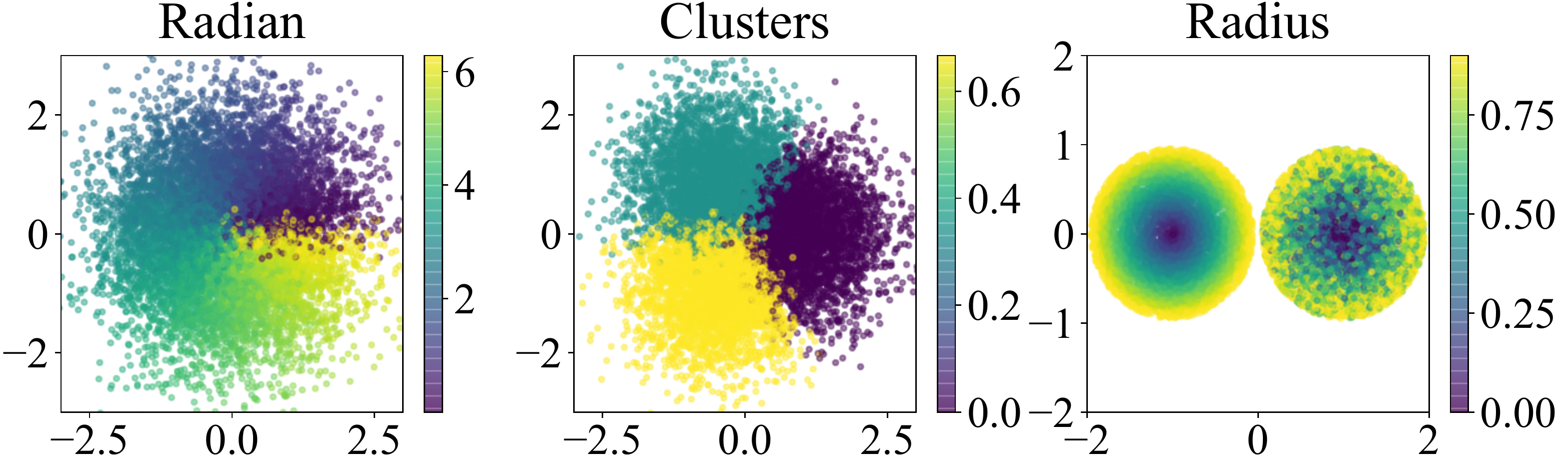}
    \caption{Toy data distributions, two axes correspond to the value of $\bx$,
        the value of $y$ is denoted by color.}
    \label{fig:data_dist}
\end{figure}

Notice that simply adding random noises to $\bx$ or $\by$ uniformly could not capture the characteristic of robust design problems.
For example, if there are two parameters $\bx_1$ and $\bx_2$, the target is $\yt$, and both parameters can achieve $\yt$, that is,
$g(\bx_1)=\yt$, $g(\bx_2)=\yt$. After adding a random noise to $\bx_1$ and $\bx_2$ uniformly, or adding a random noise to $\yt$,
the parameters $\bx_1$ and $\bx_2$ may still be equally bad considering robustness.
This indicates that the noise level should be dependent on the value of $\bx$ or $\yt$
in order to benchmark robust inverse design methods reasonably.

We define the stochastic problem with $\bx$ dependent noise:
\begin{equation}
    g_{n_x}(\bx) = g(\bx + \mathcal{n}_x(\bx, \be))
\end{equation}
A stochastic problem with $\by$ dependent noise is defined by
\begin{equation}
    g_{n_y}(\bx) = g(\bx) + \mathcal{n}_y(g(\bx), \be)
\end{equation}
A stochastic problem with $\bx$ dependent noise as well as $\by$ dependent noise is defined by
\begin{align}
    g_{n_{xy}}(\bx) & = \tilde{\by} + \mathcal{n}_y(\tilde{\by}, \be) \\
    \tilde{\by}     & = g(\bx + \mathcal{n}_x(\bx, \be))
\end{align}
Where $\epsilon$ is a random noise vector sampled from a normal distribution.
The noise functions $\mathcal{n}_x$, $\mathcal{n}_y$ and $\mathcal{n}_{xy}$ are constructed to
emphasize the parameter selection power of robust design methods for each task.
For example, the $\mathcal{n}_x$ in the Kinematics task adds smaller noise when the arm
is higher than horizon, thus the robust design method should learn to prefer to push the arm higher.

\subsection{Evaluation Metric}

We assume the training and testing datasets are IID samples from $p(\bx, \by)$.
With the objective of a robust design defined in \refeq{rid_objective},
the expected loss of an inverse design model (IDM) can be defined as:
\begin{equation}\label{resim_err}
    \bE_{\yt \sim p(\by), \bx' \sim \text{IDM}(\yt)} \mathbcal{L}_{p(\by|\bx)}(\bx', \yt)
\end{equation}
Given the testing dataset $D_t = \{y_1, y_2,...y_{n_t}\}$,
we estimate \refeq{resim_err} by:
\begin{equation}
    \frac{1}{n_t}\sum_{i=1}^{n_t} (\by_i - g_n(\text{IDM}(\by_i)))^2
\end{equation}
The standard deviation used in t-test is calculated accordingly.

\begin{figure*}
    \centering
    \includegraphics[width=\linewidth]{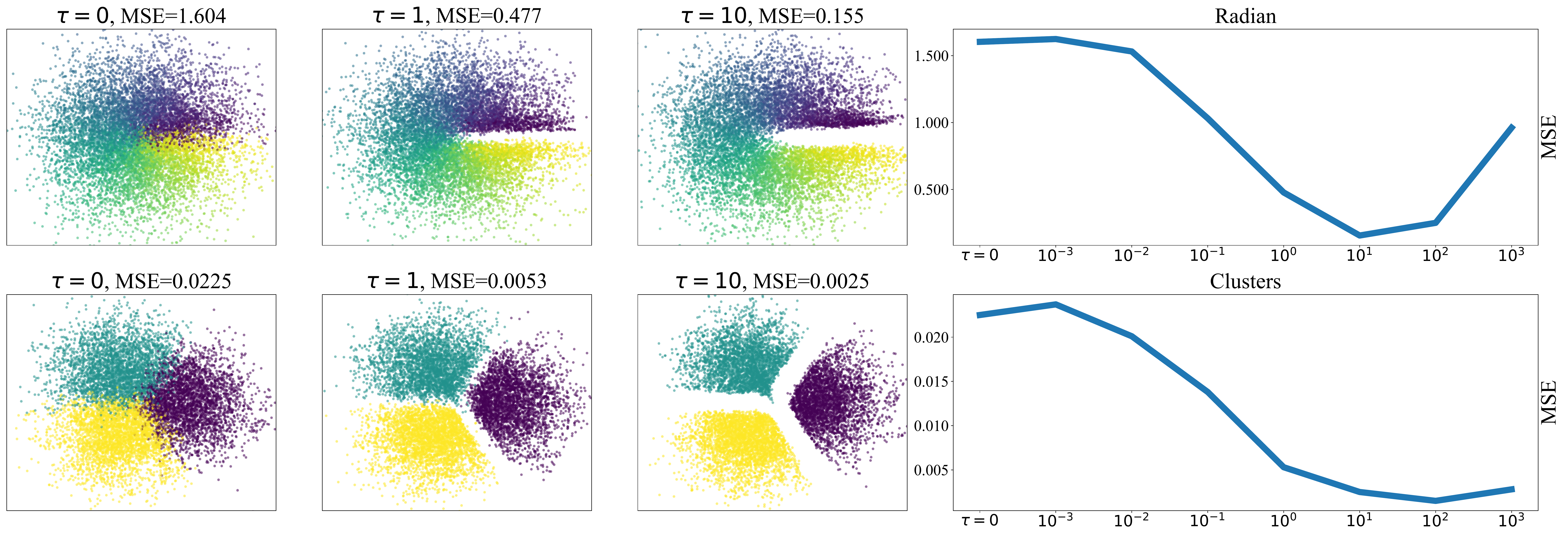}
    \caption{Effect of varying $\tau$.
        Intuitively, a larger $\tau$ will enlarge the discrepancy of weights,
        thus the model will be further away from the unstable regions.
        The upper panel shows the Radian task.
        We can see that the larger the $\tau$ is, the larger the gap near $\by=0$ is.
        We can also infer that a larger $\tau$ may not be always better, an extremely large $\tau$ may harm performance.
        The lower panel shows the Clusters task.
        We can see that the larger the $\tau$ is, the larger the gap between clusters is.
        Notice that the gap at the bottom right is larger than the other two gaps.
        Since the cost of mistake is $||\frac{2}{3}-0||_2^2$ and $||\frac{1}{3}-0||_2^2$ correspondingly.
        This phenomenon indicates a significant difference between the inverse cluster task and a robust classification task:
        in the inverse cluster task, the similarity of the label $\by$ is measured by their $l2$ distance,
        while in a classification task the distance between labels may not be defined.
        The right side shows how $\tau$ influences MSE.}
    \label{fig:cart_var_tau}
\end{figure*}

\subsection{Benchmark Performance: Q1}

We report the performance on the aforementioned three benchmark tasks in \reftab{tab:benchmark}.
We can see that our \dab{} method performs best on all the three tasks with various noise types.
We can also infer some information about how environment noises affect different inverse design methods.
First, deterministic method NA and Tandem is much more sensitive to noise,
especially on Kinematics and Balllistics tasks.
Since deterministic method will try to fit all the data points,
including the points that are affected by noise heavily.
Second, none of the methods except \dab{} exhibits the advantage on all the tasks. For example,
cINN performs well only on Kinematics; cVAE and cGAN perform best on Balllistics;
NA performs best on Meta Meterial.
Third, the performance of \dab{} is close to NA on the Meta Material task,
especially on the $\by$ dependent noise setting, the difference is not significant.
This is not a surprising result, since most spectrums in the Meta Material task are generally
smooth and not very sensitive to noise.
NA may successfully learn to smooth the $\by$ dependent noise to its mean, which has the same effect with \dab{},
thus leading to similar results.

\subsection{Visualize on Toy Problems: Q2}

We further propose three toy tasks to show when the robust inverse design problem is crucial
and why the proposed \dab{} method works. All of the toy tasks have two dimensional input $\bx$,
and a one dimensional output $y$.

\begin{itemize}
    \item[--] \textbf{Inverse Radian}:
        $\bx=(x_1, x_2)$ is drawn from independent normal distributions, the $y$ value is
        the radian value of $\bx$ in polar coordinate system.
        A gaussian noise is added to $\bx$ to construct a stochastic environment.

    \item[--] \textbf{Inverse Clusters}:
        $\bx=(x_1, x_2)$ is drawn from a gaussian mixture with three clusters, the $y$ values for different
        clusters are $0, \frac{1}{3}, \frac{2}{3}$ accordingly.
        A gaussian noise is added to $\bx$ to construct a stochastic environment.

    \item[--] \textbf{Inverse Radius}:
        $\bx=(x_1, x_2)$ is drawn from two spheres with centers at $(0, -1)$ and $(0, 1)$ respectively,
        the $y$ value is the distance to the corresponding center.
        We add a gaussian noise to $\by$ on the right cluster, while leave the left cluster clean.
\end{itemize}

The training data distribution of the toy tasks are visualized in \reffig{fig:data_dist}.
The visualization of the toy tasks
show why \dab{} works from different perspectives.
The Inverse Radian task is visualized in \reffig{fig:intro_cart},
this task clearly shows that \dab{} could learn to avoid the regions where the value of $\by$ is most sensitive to noise.
The Inverse Clusters task is visualized in \reffig{fig:cart_var_tau},
this task reflects the relationship and difference between robust classification and robust inverse design.
They are similar since they both try to separate different clusters further apart.
But, the force that pushes different cluster apart is influenced by their $\by$ value
in robust inverse design, while robust classification treats each class equally.
The Inverse Radius task visualized in \reffig{fig:two_cluster}
shows that when two different parameters $\bx_1$ and $\bx_2$ both achieve $\yt$,
\dab{} can learn to select the parameter that is more robust in $\by$ value.

\subsection{The Effect of $\tau$ in \dab{}: Q3}

To explore the effect of $\tau$, we conduct experiments with different $\tau$ as reported in \reffig{fig:cart_var_tau}.
From the visualization, we can clearly see how $\tau$ influences the conditional distribution
learned by the generative model.
Specifically, the larger the value of $\tau$ is, the more conservative the model becomes.
This indicates that the learned model can ignore more data points which are believed to be less robust.
However, if $\tau$ is too large that too many training data points are ignored, the MSE starts to increase.

\section{Conclusion}

We propose a \dab{} method to solve the robust inverse design problem, 
which aims at generating design parameters that are insensitive to noise with knowledge learned from a dataset.
To evaluate the performance of robust inverse design methods,
we propose three robust design benchmarks that adopted from real-world problems.
The experiments on the benchmark problems exhibit
that our method can learn robustness from a dataset efficiently.
We further conduct several experiments to visually show how \dab{} works.
The results indicate that \dab{} can learn to ignore the data points that are most harmful to robustness.
Besides, \dab{} can successfully learn the training data distribution,
which contains expert knowledge used to explore the feasible parameter manifolds.

\section{Acknowledgments}
 We thank Han-Jia Ye and Yang Yang for their insightful discussion.

\bibliography{biblio.bib}

\end{document}